\documentclass[journal,two side,web]{ieeecolor}
\usepackage{multirow}
\usepackage{etoolbox}
\makeatletter
\@ifundefined{color@begingroup}%
{\let\color@begingroup\relax
\let\color@endgroup\relax}{}%
\def\fix@ieeecolor@hbox#1{%
\hbox{\color@begingroup#1\color@endgroup}}
\patchcmd\@makecaption{\hbox}{\fix@ieeecolor@hbox}{}{\FAILED}
\patchcmd\@makecaption{\hbox}{\fix@ieeecolor@hbox}{}{\FAILED}
\usepackage{tmi}
\usepackage{cite}
\usepackage{amsmath,amssymb,amsfonts}
\usepackage{algorithmic}
\usepackage{graphicx}
\usepackage{textcomp}
\usepackage{arydshln}
\usepackage{booktabs}
\def\BibTeX{{\rm B\kern-.05em{\sc i\kern-.025em b}\kern-.08em
    T\kern-.1667em\lower.7ex\hbox{E}\kern-.125emX}}
\begin{document}
\title{LapaFM: A Laparoscopic Foundation Model via Multi-Stage Continual Learning}
\author{Xiaobo Zhu, Shiwei Wu, Yu An, \IEEEmembership{Member, IEEE}, Yuechen Tao, He Sun, Song Zhang, Jianbo Wang, Xiaojing Wang, Junwei Liu, Wei Guo, Yanfei Shao, Xuan Zhao, Jiangwei Xiao, Wenbo Niu, Bo Wei, Yuhan Wang, Gang Hu, Fei Liu, Kunshan He, Chongwei Chi, Jianqiang Tang, Kang Zhang, Jie Tian, \IEEEmembership{Fellow, IEEE}, and Zhenyu Liu, \IEEEmembership{Member, IEEE}
\thanks{This work was supported by the National Key Research and Development Program of China grant 2024YFF1207400 and 2023YFC3402800, the National Natural Science Foundation of China grant 62333022, 92359302, and 92259301, the Beijing Natural Science Foundation grant JQ23034 and L232097, and the Medical and Health Science and Technology Program of Zhejiang Province (China) under Grant No. 2023RC309. (Xiaobo Zhu, Shiwei Wu, and Yu An contribute equally to the article. Corresponding authors: Kang Zhang, Jie Tian, and Zhenyu Liu.)}
\thanks{Xiaobo Zhu, Yuechen Tao, Song Zhang, Chongwei Chi, Kunshan He, Jie Tian, and Zhenyu Liu are with the CAS Key Laboratory of Molecular Imaging and Beijing Key Laboratory of Molecular Imaging, Institute of Automation, Chinese Academy of Sciences, Beijing 100190, China, and also with the School of Artificial Intelligence, University of Chinese Academy of Sciences, Beijing 100049, China (e-mail: zhuxiaobo2023@ia.ac.cn; taoyuechen2023@ia.ac.cn; zhangsong2021@ia.ac.cn; chongwei.chi@digipmc.com; kunshan.he@digipmc.com; jie.tian@ia.ac.cn; zhenyu.liu@ia.ac.cn).}
\thanks{Shiwei Wu is with School of Computer Science and Engineering, Nanjing University of Science and Technology, Nanjing, Jiangsu Province, 210094, China (e-mail: wushiwei@njust.edu.cn).}
\thanks{Yu An and He Sun are with School of Engineering Medicine, Beihang University, Beijing, 100191, China, and also with Key Laboratory of Big Data-Based Precision Medicine (Beihang University), Ministry of Industry and Information Technology of China, Beijing, 100191, China (e-mail: yuan1989@buaa.edu.cn; hesun@buaa.edu.cn).}
\thanks{Jianbo Wang, Xiaojing Wang, and Junwei Liu are with Department of Gastroenterology, Lishui Central Hospital, The Fifth Affiliated Hospital of Wenzhou Medical University, Lishui, Zhejiang Province, 323020, China (e-mail: wansao723@163.com; wxjing1986@163.com; yixiliu120@163.com).}
\thanks{Wei Guo is with Heping Hospital Affiliated to Changzhi Medical College, Changzhi, Shanxi Province, 425599, China (e-mail: uoei2@163.com).}
\thanks{Yanfei Shao and Xuan Zhao are with Shanghai Jiao Tong University School of Medicine, Ruijin Hospital, Department of General Surgery, Shanghai, 200025, China (e-mail: syf01p17@rjh.com.cn; zx12449@rjh.com.cn). }
\thanks{Jiangwei Xiao is with Department of Gastrointestinal Surgery, Clinical Medical College and The First Affiliated Hospital of Chengdu Medical College, Chengdu, Sichuan Province, 610500, China (e-mail: xiaojiangwei@126.com).}
\thanks{Wenbo Niu is with The Fourth Hospital of Hebei Medical University, Shijiazhuang, Hebei Province, 050011, China (e-mail: 39728149@qq.com).}
\thanks{Bo Wei is with Department of Gastrointestinal Surgery, the Third Affiliated Hospital of Sun Yat-sen University, Guangzhou 510630, China (e-mail: weibo3@mail.sysu. edu.cn).}
\thanks{Yuhan Wang, Gang Hu, and Jianqiang Tang are with the Department of Colorectal Surgery, National Cancer Center/National Clinical Research Center for Cancer/Cancer Hospital, Chinese Academy of Medical Sciences and Peking Union Medical College, Beijing, 100021, China (e-mail: wyh06112023@163.com; doctorhuhu@126.com; doc\_tjq@hotmail.com).}
\thanks{Fei Liu and KangZhang are with State Key Laboratory for Eye Health, Eye Hospital, Wenzhou Medical University, Wenzhou, Zhejiang Province, 325027, China (e-mail: liufei\_2359@163.com; kang.zhang@gmail.com).}}

\maketitle

\begin{abstract}
Artificial intelligence–based analysis of intraoperative laparoscopic videos has the potential to improve surgical decision-making and precision. However, effective surgical assistance requires models that can generalize across diverse intraoperative tasks. Here we present LapaFM, a foundation model for laparoscopic surgery trained on more than 20,000 minutes of surgical videos using a multi-stage continual learning strategy that mitigates catastrophic forgetting and inter-task interference. LapaFM learns multimodal visual representations and supports multiple tasks, including smoke removal, surgical phase recognition, instrument and polyp segmentation, and lesion identification. The model was evaluated on 13 retrospective and two prospective cohorts collected from different imaging devices. LapaFM consistently outperformed task-specific models and existing foundation models across datasets, and also generalized to gastrointestinal endoscopy tasks, including polyp detection and classification. Notably, it recovered over 90\% of smoke-obscured frames and achieved clinician-level or superior performance in lymph node metastasis prediction and prospective polyp classification.
\end{abstract}

\begin{IEEEkeywords}
Foundation model, laparoscopic surgery, multi-task assistance, intraoperative imaging, Continual Learning.
\end{IEEEkeywords}
\section{Introduction}
\label{sec:introduction}

\IEEEPARstart{M}{inimally} invasive laparoscopic surgery, performed through small incisions (0.5–1.5 cm) using a laparoscope, has become the standard in many clinical domains due to reduced trauma, faster recovery, and fewer complications\cite{bonjer2015randomized}. Modern procedures generate large volumes of multimodal intraoperative video data, including white-light, fluorescence\cite{wang2023fluorescence}, and angiographic imaging, capturing rich information on anatomical structures, surgical workflows, and instrument--tissue interactions. However, despite their clinical value, these data remain underutilized due to the lack of effective analytical frameworks.

Artificial intelligence (AI) has shown promise in medical image analysis, yet its application in laparoscopic surgery remains limited by poor generalization, dependence on large labeled datasets\cite{feng2022development}, and insufficient modeling of complex intraoperative dynamics. Recently, foundation models (FMs), trained on large-scale unlabeled data, have demonstrated strong capability in learning transferable representations across diverse medical modalities\cite{moor2023foundation}. While such models have achieved success in CT, MRI, ultrasound, and pathology, there is still a lack of foundation models specifically designed for laparoscopic intraoperative multimodal imaging analysis.

Laparoscopic surgery presents several key challenges, including visual obstruction from surgical smoke and tissue occlusion, imprecise localization of instruments and anatomical structures, and difficulty in distinguishing lesions from surrounding tissues. Existing AI approaches are predominantly task-specific, limiting their ability to capture the complex and dynamic nature of surgical scenes and to generalize across multiple intraoperative tasks. A unified foundation model tailored for laparoscopic imaging has the potential to address these limitations by providing robust, task-agnostic representations.

To this end, we propose LapaFM, a multimodal intraoperative imaging foundation model for laparoscopic surgery. LapaFM is trained on over 20,000 minutes of multimodal surgical videos using a multi-stage continual learning strategy that integrates white-light imaging (WLI) and pseudo-color imaging (PCI) while mitigating catastrophic forgetting. The model supports a wide range of intraoperative tasks, including smoke removal, surgical phase recognition, instrument and polyp segmentation, polyp detection and classification, and lymph node metastasis prediction, achieving strong performance across all tasks. Extensive evaluations on datasets from multiple devices demonstrate robust generalization, and prospective validation (NCT06846229) further confirms its clinical applicability. The main contributions of this work are as follows:
\begin{itemize}
\item Construction of a large-scale multimodal laparoscopic dataset comprising over 20,000 minutes of surgical videos, covering both white-light and pseudo-color imaging modalities.
\item Development of a multi-stage continual learning framework to progressively integrate multimodal information while preserving modality-specific knowledge, thereby alleviating interference and catastrophic forgetting.
\item Establishment of a unified foundation model for laparoscopic surgery, demonstrating state-of-the-art performance across multiple intraoperative tasks with strong cross-domain generalization.
\end{itemize}

\begin{figure*}[ht]
\centering
\includegraphics[width=1\textwidth]{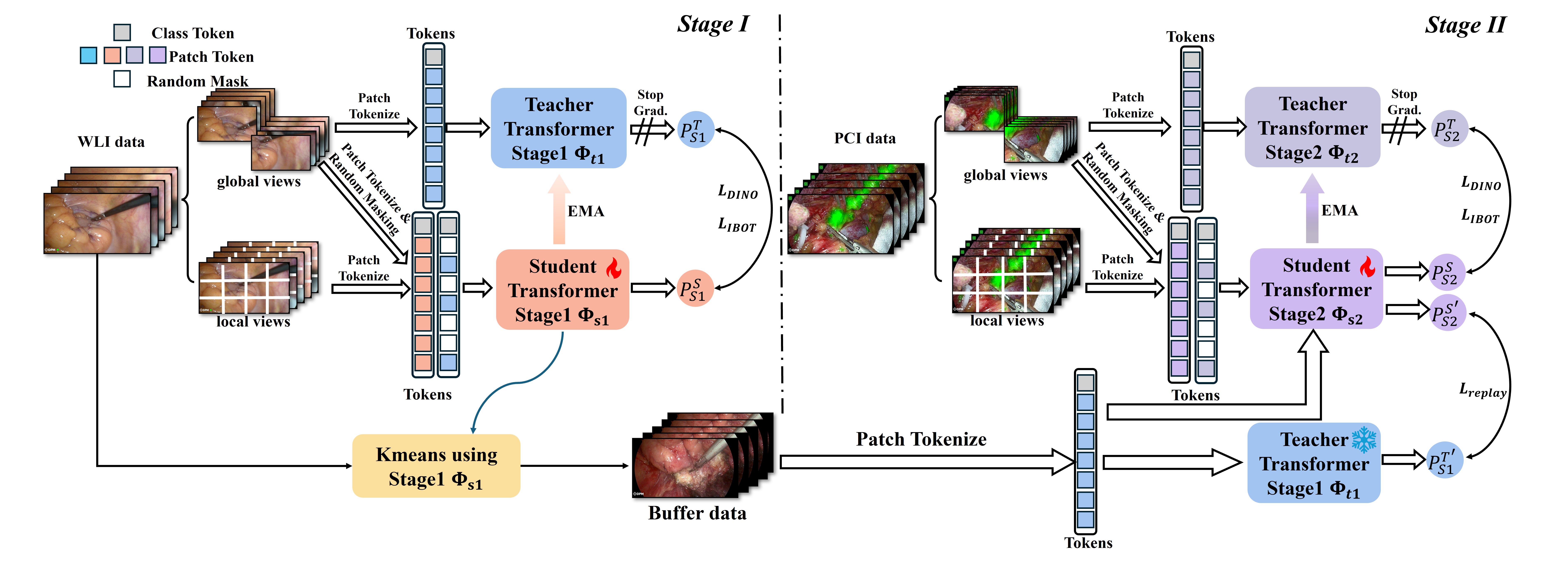}
\caption{A multi-stage continual learning strategy is employed during pretraining. In Stage I, the model learns white-light imaging (WLI) representations. In Stage II, pseudo-color imaging (PCI) features are introduced while reviewing white-light features via a frozen teacher network to mitigate modality conflict and catastrophic forgetting.}
\label{fig: CL_Framework}
\end{figure*}
\section{METHODOLOGY}
\subsection{Multi-stage Continual Learning Pretraining}

We propose a multi-stage continual self-supervised learning framework for cross-domain intraoperative imaging representation learning, as shown in Fig.\ref{fig: CL_Framework}. Given sequentially observed datasets from different imaging modalities, denoted as $\mathcal{D}_1$ (e.g., white-light imaging, WLI) and $\mathcal{D}_2$ (e.g., pseudo-color imaging, PCI), the objective is to learn transferable visual representations while mitigating catastrophic forgetting. The proposed framework follows a teacher–student paradigm based on Vision Transformers (ViTs) and consists of two sequential training stages, namely an initial self-supervised pretraining stage and a continual adaptation stage with cross-stage knowledge preservation.

In Stage I, self-supervised pretraining is performed on $\mathcal{D}_1$. For each image $x$, multi-view augmentation generates global and local views, which are tokenized as
\begin{equation}
\mathbf{z} = \mathrm{PatchEmbed}(v), \quad 
\tilde{\mathbf{z}} = [\mathbf{z}_{cls}; \mathbf{z}_1,\dots,\mathbf{z}_N].
\end{equation}
A masking operator $\mathcal{M}(\cdot)$ is applied to the student input:
\begin{equation}
\tilde{\mathbf{z}}^{mask} = \mathcal{M}(\tilde{\mathbf{z}}),
\end{equation}
which enables masked token modeling. The masked and unmasked tokens are fed into the student and teacher encoders, respectively:
\begin{equation}
p_{S_1}^{S} = f_{\theta_s}(\tilde{\mathbf{z}}^{mask}), \quad
p_{S_1}^{T} = f_{\theta_t}(\tilde{\mathbf{z}}).
\end{equation}
The teacher parameters are updated via exponential moving average (EMA):
\begin{equation}
\theta_t \leftarrow m \theta_t + (1 - m)\theta_s,
\end{equation}
where $m$ denotes the momentum coefficient. To learn robust representations, we combine global self-distillation and local masked prediction objectives. Global and local consistency are enforced using
\begin{align}
\mathcal{L}_{\mathrm{DINO}} = - \sum_{i} p^{T}_{S_i} \log p^{S}_{S_i}, \quad \\
\mathcal{L}_{\mathrm{IBOT}} = - \sum_{j \in \mathcal{M}} p^{T,j}_{S_i} \log p^{S,j}_{S_i}.
\end{align} 
The overall objective in Stage I is:
\begin{equation}
\mathcal{L}_{S1} = \mathcal{L}_{\mathrm{DINO}} + \lambda_1 \mathcal{L}_{\mathrm{IBOT}}.
\end{equation}
After convergence, features are extracted from the white-light pretraining data using the trained student encoder. K-means clustering is then applied to identify the top 10\% of samples with the most informative feature representations, which are selected as buffer data:
\begin{equation}
\mathcal{B} = \mathrm{Top}_{10\%} \left( \mathrm{KMeans}\big(f_{\theta_s}(x)\big) \right), \quad x \in \mathcal{D}_1.
\end{equation}

In Stage II, the model is adapted to $\mathcal{D}_2$ while preserving knowledge from $\mathcal{D}_1$. A dual-teacher scheme is employed, consisting of a trainable teacher $f_{\theta_t^{(2)}}$ and a frozen teacher $f_{\theta_t^{(1)}}$. For $x \in \mathcal{D}_2$ and $x' \in \mathcal{B}$,
\begin{equation}
p^S_{S_2} = f_{\theta_s}(x), \quad p^{S'}_{S_2} = f_{\theta_s}(x'),
\end{equation}
\begin{equation}
p^T_{S_2} = f_{\theta_t^{(2)}}(x), \quad p^{T'}_{S_1} = f_{\theta_t^{(1)}}(x').
\end{equation}

The objective combines self-supervised learning and memory replay:
\begin{equation}
\mathcal{L}_{\mathrm{SSL}} = \mathcal{L}_{\mathrm{DINO}} + \lambda_2 \mathcal{L}_{\mathrm{IBOT}},
\end{equation}
\begin{equation}
\mathcal{L}_{\mathrm{replay}} = \left\| p^{S'}_{S_2} - p^{T'}_{S_1} \right\|_2^2,
\end{equation}
\begin{equation}
\mathcal{L}_{S2} = \mathcal{L}_{\mathrm{SSL}} + \lambda_3 \mathcal{L}_{\mathrm{replay}}.
\end{equation}

By integrating multi-view self-supervision, prototype-based memory replay, and cross-stage knowledge distillation into a unified transformer framework, the proposed method enables effective continual adaptation across imaging modalities while preserving previously learned representations.

\begin{figure*}[!ht]
\centering
\includegraphics[width=1\textwidth]{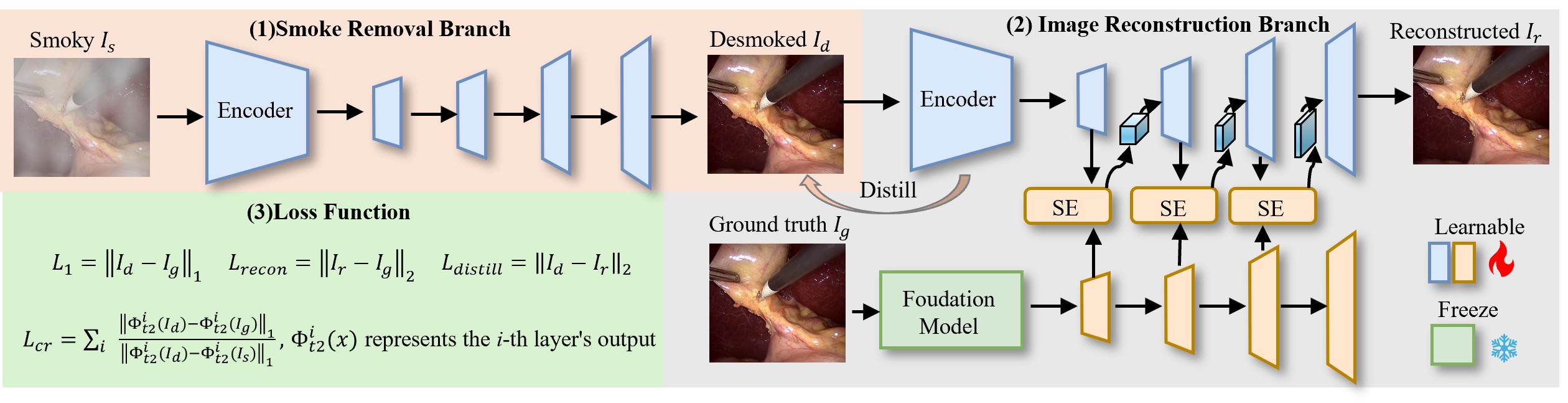}
\caption{Overview of the proposed smoke removal model. During training, the image reconstruction branch refines the output of the smoke removal branch by leveraging semantic priors extracted from the ground truth via LapaFM. A semantic embedding module is employed to guide high-quality reconstruction. During inference, only the smoke removal branch is retained, while the pretrained prior knowledge from LapaFM is transferred under the supervision of the distillation loss $\mathcal{L}_{\mathrm{distill}}$.}
\label{fig: desmoking_framework}
\end{figure*}
\subsection{Downstream De-Smoking via Dual-Network Semantic Distillation}

To address the laparoscopic smoke removal problem, we propose a dual-network semantic distillation framework that transfers semantic priors from LapaFM to a lightweight de-smoke network, as illustrated in Fig.\ref{fig: desmoking_framework}. The framework consists of a smoke removal branch and an image reconstruction branch, which are jointly optimized during training to enhance detail recovery in smoke-occluded regions while maintaining real-time performance.

\subsubsection{Smoke Removal Branch}
{Given a smoky laparoscopic image $I_s \in \mathbb{R}^{H \times W \times 3}$, the smoke removal branch aims to produce an initial de-smoked image:
\begin{equation}
I_d = F_{\theta_d}(I_s),
\end{equation}
where $F_{\theta_d}$ denotes a lightweight encoder–decoder network (Dehazeformer-B\cite{song2023vision}). This branch primarily focuses on removing smoke artifacts and restoring coarse structures. The supervision is provided by the ground-truth clean image $I_g$, and the corresponding reconstruction loss is defined as:
\begin{equation}
\mathcal{L}_{1} = \| I_d - I_g \|_1.
\end{equation}
}
\subsubsection{Image Reconstruction Branch}
{To further enhance fine details and semantic consistency, we introduce a reconstruction subnetwork:
\begin{equation}
I_r = F_{\theta_r}(I_d),
\end{equation}
where $F_{\theta_r}$ shares a similar backbone but is augmented with Semantic Embedding (SE) modules. The SE modules inject semantic priors extracted from the pretrained foundation model (LapaFM) $F_{\theta_f}$:
\begin{equation}
\mathbf{s}_i = F_{\theta_f}^{(i)}(I_g),
\end{equation}
where $\mathbf{s}_i$ denotes the feature map from the $i$-th layer of the frozen foundation model. These semantic features are fused into the reconstruction network via SE modules to guide detail restoration, especially in heavily occluded regions.

The reconstruction branch is supervised by:
\begin{equation}
\mathcal{L}_{recon} = \| I_r - I_g \|_2^2.
\end{equation}
}
\subsubsection{Semantic Distillation}
{To transfer semantic knowledge from the pretrained model to the lightweight network, we introduce a feature-level distillation loss. Specifically, we align intermediate representations between the reconstruction branch and the foundation model:
\begin{equation}
\mathcal{L}_{cr} = \sum_i \frac{\left\|\Phi_{t 2}^i\left(I_d\right)-\Phi_{t 2}^i\left(I_g\right)\right\|_1}{\left\|\Phi_{t 2}^i\left(I_d\right)-\Phi_{t 2}^i\left(I_s\right)\right\|_1}
\end{equation}
where $\Phi_{t 2}^i(\cdot)$ represents the $i$-th layer's output.

In addition, a distillation constraint is imposed between the outputs of the two branches:
\begin{equation}
\mathcal{L}_{distill} = \| I_d - I_r \|_2^2,
\end{equation}
which encourages the smoke removal branch to inherit refined structural information from the reconstruction branch.
}

The total training loss is defined as:
\begin{equation}
\mathcal{L} =
\alpha_1 \mathcal{L}_{1}
+ \alpha_2 \mathcal{L}_{recon}
+ \alpha_3 \mathcal{L}_{distill}
+ \alpha_4 \mathcal{L}_{cr},
\end{equation}
where $\alpha_1, \alpha_2, \alpha_3$, and $\alpha_4$ are weighting coefficients.

During inference, only the smoke removal branch $F_{\theta_d}$ is retained, while the reconstruction branch and foundation model are discarded. Through semantic distillation, the lightweight model implicitly encodes high-level semantic priors, enabling efficient and high-quality smoke removal suitable.

\section{EXPERIMENTAL SETTINGS}
\subsection{Pretraining Data}
For pretraining, we collaborated with DPM to collect approximately 250 laparoscopic surgical videos from five clinical centers, including Heping Hospital Affiliated to Changzhi Medical College, Shanghai Ruijin Hospital, The First Affiliated Hospital of Chengdu Medical College, The Fourth Hospital of Hebei Medical University, and The Third Affiliated Hospital of Sun Yat-sen University. All videos were acquired using a DPM self-developed 3D endoscopic fluorescence imaging system certified by NMPA, FDA, and CE. The dataset includes two intraoperative imaging modalities, namely white-light and pseudo-color frames. To ensure data quality, noisy and irrelevant segments (e.g., external scenes and blurred frames) were removed, followed by modality-wise separation. Uniform frame sampling at 2 fps was applied to reduce redundancy while preserving key spatiotemporal information. The final dataset comprises 12,097 minutes of white-light video and 1,577 minutes of pseudo-color video, corresponding to 1,451,670 and 189,280 frames, respectively.

\subsection{Downstream Datasets}
The proposed method is evaluated on multiple downstream tasks using both public and in-house clinical datasets, including classification, segmentation, surgical phase recognition, and smoke removal. For polyp detection, the CP-CHILD dataset\cite{wang2020improved} is used, containing approximately 9,500 endoscopic images; the two subsets are merged into 8,100 training and 1,400 testing images. A prospective dataset (Propolyp-cls) collected from 115 patients at Wenzhou Fifth Affiliated Hospital comprises 300 polyp images (217 adenomas and 83 hyperplastic polyps). For prospective detection (Propolyp-detection), these images are combined with 100 non-polyp frames randomly sampled from colonoscopy videos without preprocessing, introducing realistic noise and motion artifacts.

For gastrointestinal disease classification, the WCE-CCD\cite{montalbo2022diagnosing} dataset is adopted, including four categories with 3,200 training, 2,000 validation, and 800 testing images. For lymph node metastasis prediction, the CRCLNM\cite{zhu2024prediction} dataset is used, where white-light and pseudo-color modalities are selected, with splits of 678/281/280 for training/validation/testing.

For segmentation, EndoVis2017\cite{allan20192017} and EndoVis2018\cite{allan20202018} are used for surgical instrument segmentation (750/300 and 2,235/997 training/testing images, respectively), and polyp segmentation follows the PraNet\cite{fan2020pranet} protocol with 1,450 training images and four test sets (CVC-ClinicDB, CVC-ColonDB, EndoScene, and ETIS).

For surgical phase recognition, the AutoLaparo\cite{wang2022autolaparo} and Cholec80\cite{twinanda2016endonet} datasets are used, with standard splits (10/4/7 and 40/40 videos, respectively), and all videos are sampled at 1 fps.

For smoke removal, both synthetic and real datasets are employed. The synthetic dataset (DS$_{\text{Synthetic}}$) contains 15,000 smoke-free images and corresponding synthetic smoke images, while the real dataset (DS$_{\text{Real}}$)\cite{xia2025vivo} includes 3,464 paired images from 109 surgeries, split into 2,424/520/520 for training/validation/testing.

These datasets collectively provide a comprehensive evaluation across diverse modalities and clinical scenarios.
\begin{table*}[!htbp]
\renewcommand\arraystretch{1.3}
\setlength\tabcolsep{10pt} 
\caption{Quantitative comparison of LapaFM with seven state-of-the-art desmoking/dehazing models or general image restoration models—AECR-Net, Ds-CycleGAN, Cyclic-DeGAN,DEA-Net, MB-Taylor-B V2, SGDN, and MambaIRv2-S—on both synthetic and real laparoscopic datasets. LapaFM achieves superior performance across all metrics}
\centering
\begin{tabular}{ccccccc}
\hline
\multirow{2}{*}{\textbf{Model}} & \multicolumn{3}{c}{\textbf{DS\_Synthetic}} & \multicolumn{3}{c}{\textbf{DS\_Real}}    \\ \cline{2-7} 
                       & \textbf{SSIM$\uparrow$}    & \textbf{PSNR$\uparrow$ }    &\textbf{ CIEDE-2000$\downarrow$ }  & \textbf{SSIM$\uparrow$}    & \textbf{PSNR$\uparrow$}     & \textbf{CIEDE-2000$\downarrow$} \\ \hline
AECR-Net (CVPR'21) \cite{wu2021contrastive}               & 0.949   & 27.487   & 3.696        & 0.870  & 25.751  & 4.859     \\
DS-CycleGAN (TCBB'22)\cite{zhou2022synchronizing}            & 0.792   & 20.412   & 9.495        & 0.766  & 21.927  & 7.688     \\
Cyclic-DeGAN (CBM'20) \cite{venkatesh2020unsupervised}           & 0.865   & 21.280    & 7.525        & 0.833   & 22.397   & 7.173     \\
DEA-Net (TIP'24) \cite{chen2024dea}                & 0.964   & 29.706   & 2.833       & 0.846 & 25.289 & 5.029    \\
MB-Taylor-B V2 (TPAMI'25) \cite{jin2025mb}         & 0.969   & 29.551   & 2.658        & 0.871 & 25.883  & 4.853    \\
SGDN (AAAI'25) \cite{fang2025guided}                   & 0.970    & 29.670    & 2.605        & 0.877 & 26.305 & 4.465    \\
MambaIRv2-S (CVPR'25) \cite{guo2025mambairv2}            & 0.964   & 28.730    & 2.814        & 0.869  & 25.234 & 4.891    \\
\textbf{Ours }                  & \textbf{0.972}   & \textbf{30.293}   & \textbf{2.517}        & \textbf{0.881}   & \textbf{26.686}   & \textbf{4.275}      \\ \hline
\label{tab:1}
\end{tabular}
\end{table*}

\subsection{Implementation Details}
\label{subsec: Implementation Details}
For surgical phase recognition, polyp detection, and classification, a linear classification head is appended to the 1024-dimensional class token of LapaFM, with only the head trained while the backbone remains frozen. For lymph node metastasis prediction, features are extracted from both white-light and pseudo-color images using the frozen model, and their class tokens are concatenated for classification. Performance is evaluated using ACC, AUC, and F1 score. For surgical instrument and polyp segmentation, a three-layer inverse convolutional pyramid head is applied to patch tokens, progressively upsampling feature maps to restore the original resolution (518×518), with only the segmentation head trained. Segmentation is evaluated using IoU and Dice. LapaFM is pretrained for 100 epochs over 10 days on four NVIDIA A100 GPUs using PyTorch 2.4, with a base learning rate of $1\times10^{-6}$ and weight decay of 0.04. Multi-view augmentation generates two global and eight local views per sample, with carefully designed cropping strategies to ensure spatial diversity, while temperature scaling and masking ratios are empirically tuned to stabilize training and enhance representation learning.

\begin{figure*}[!ht]
\centering
\includegraphics[width=\textwidth]{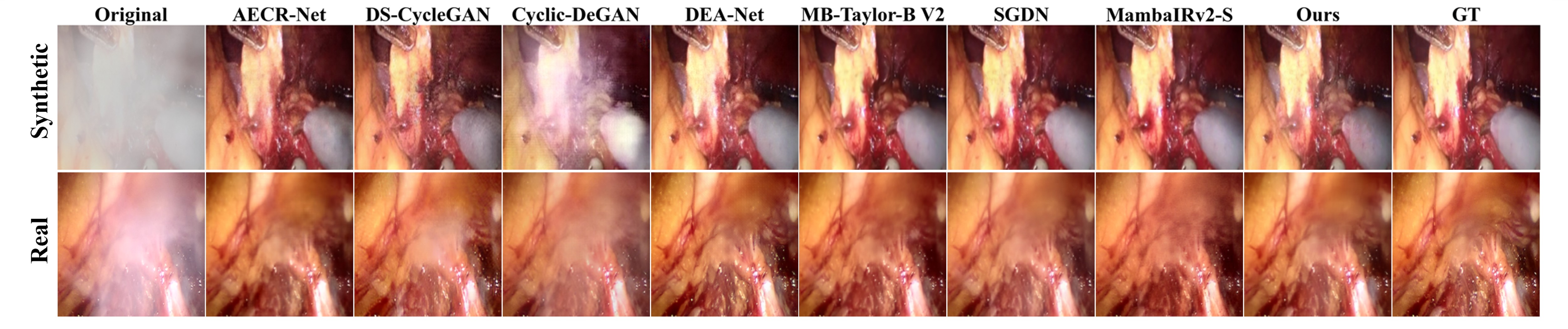}
\caption{Laparoscopic smoke removal visualization. Qualitative results demonstrate that LapaFM effectively removes surgical smoke while preserving anatomical details, outperforming other methods on both synthetic and real data.}
\label{fig: desmoking_visualization}
\end{figure*}
\section{RESULTS}
\subsection{Model performance on laparoscopic desmoking task}
Smoke generated during laparoscopic surgery can obscure tissue structures and degrade image quality, complicating intraoperative decision-making. Improving image clarity in such scenarios remains an urgent need in clinical settings\cite{hong2023mars}. To assess the performance gain enabled by distilling our smoke removal model from LapaFM, we compared it against a range of state-of-the-art de-smoke, dehazing, and general image restoration approaches, including AECR-Net\cite{wu2021contrastive}, DS-CycleGAN\cite{zhou2022synchronizing}, Cyclic-DeGAN\cite{venkatesh2020unsupervised}, DEA-Net\cite{chen2024dea}, MB-Taylor-B V2\cite{jin2025mb}, SGDN\cite{fang2025guided}, MambaIRv2-S\cite{guo2025mambairv2}. We adopted three widely used full-reference metrics—Peak Signal-to-Noise Ratio (PSNR), Structural Similarity Index (SSIM), and CIEDE2000\cite{gaurav2005ciede2000}—to quantify restoration quality. Higher PSNR and SSIM scores reflect better structural fidelity, while lower CIEDE2000 values indicate superior perceptual quality.

As shown in Table \ref{tab:1}, our method achieved the highest quantitative performance on both synthetic and real-world datasets. On the synthetic dataset, it yielded the best SSIM and PSNR scores. Specifically, compared to the second-best model, DEA-Net, our approach improved PSNR to 30.293 dB, representing a 1.98\% gain. As illustrated in Fig.\ref{fig: desmoking_visualization}, visual comparisons further demonstrate that our model effectively eliminates synthetic smoke and accurately restores occluded structures. In addition, beyond synthetic benchmarks, our method generalizes effectively to real-world surgical videos, showing robust smoke removal and reliable recovery of obscured anatomical details.

\subsection{Surgical phase recognition and instrument segmentation}
Computer-assisted intraoperative navigation has advanced significantly in recent years, with the goal of enhancing surgical precision and efficiency across diverse procedures\cite{ahmed2024deep}. These technologies leverage automated recognition of surgical phases and accurate segmentation of instruments to support real-time decision-making, improve workflow analysis, and facilitate robotic and minimally invasive interventions.

\begin{figure}[!ht]
\centering
\includegraphics[width=0.5\textwidth]{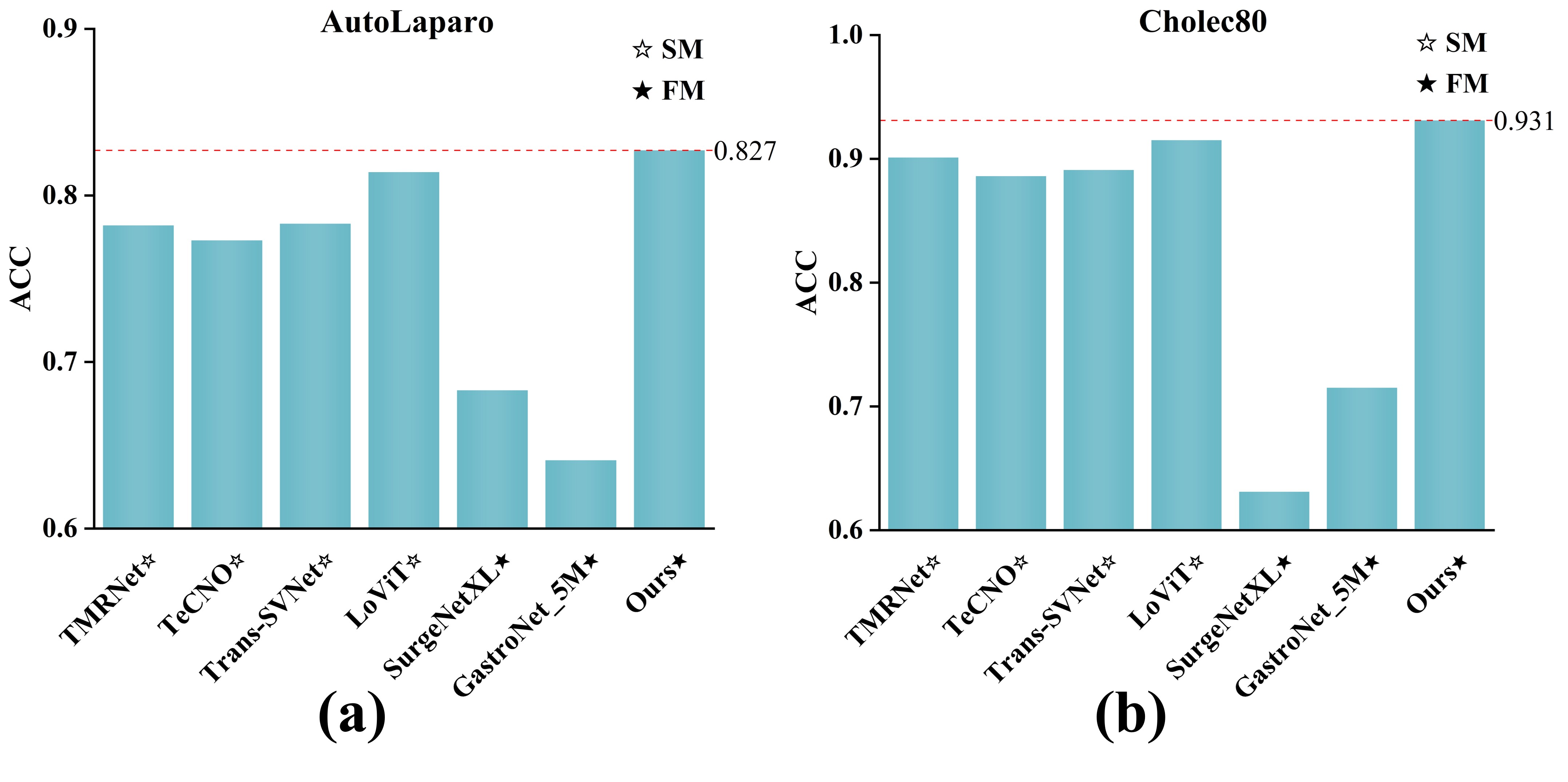}
\caption{Performance validation for surgical stage recognition task. a. Performance comparison of models on the AutoLaparo dataset for surgical phase recognition task. b. Performance comparison of models on the Cholec80 dataset for surgical phase recognition task.}
\label{fig: Surgical_phase_recognition}
\end{figure}
\subsubsection{Surgical phase recognition}
To evaluate the effectiveness of LapaFM in surgical applications, we benchmarked it against four task-specific models (SM)(TMRNet\cite{jin2021temporal}, TeCNO\cite{czempiel2020tecno}, Trans-SVNet\cite{gao2021trans}, and LoViT\cite{liu2025lovit}) and two foundation models (FM) (SurgeNetXL\cite{de2025scaling} and GastroNet\_5M\cite{jong2025gastronet}) on the AutoLaparo and Cholec80 datasets (Fig.\ref{fig: Surgical_phase_recognition}). On the AutoLaparo dataset, which is characterized by complex procedures, anatomical variability, and long temporal dependencies, most specialized models achieved accuracies around 0.78, while LoViT exceeded 0.81. LapaFM achieved the best performance (0.827), demonstrating the advantage of large-scale pretrained semantic representations in handling complex surgical workflows. On the Cholec80 dataset, all methods achieved improved performance due to its standardized workflow, with most specialized models exceeding 0.88 and LoViT approaching 0.91. LapaFM again achieved the highest accuracy (0.931), indicating strong recognition capability in structured scenarios. In contrast, SurgeNetXL and GastroNet showed substantially lower performance (approximately 0.63), suggesting limited cross-procedure generalization and reduced robustness to distribution shifts.
\begin{table}[!htbp]
\renewcommand\arraystretch{1.3}
\setlength\tabcolsep{5pt} 
\caption{Quantitative results on surgical instrument segmentation show that LapaFM outperforms three task-specific models and two general-purpose segmentation models, achieving the highest IoU and Dice scores.}
\centering
\begin{tabular}{ccccc}
\hline
{ }                         & \multicolumn{2}{c}{{ \textbf{Endovis2017}}}                        & \multicolumn{2}{c}{{ \textbf{Endovis2018}}}                        \\ \cline{2-5} 
\multirow{-2}{*}{{ \textbf{Model}}}  & { \textbf{IOU}}            & { \textbf{DICE}}           & { \textbf{IOU}}            & { \textbf{DICE}}           \\ \hline
{ SSIS-Seg (TMI'22)\cite{colleoni2022ssis}}         & { 0.867}          & { 0.926}          & { 0.613}          & { 0.661}          \\
{ Active\_learning (MIA'24)\cite{peng2024reducing}} & { 0.736}          & { 0.846}          & { 0.705}          & { 0.816}          \\
{ MCF-SMSIS (CBM'24)\cite{wu2024mcf}}        & { 0.722}          & { 0.831}          & { 0.663}          & { 0.794}          \\
{ SurgicalSAM (AAAI'24)\cite{yue2024surgicalsam}}     & { 0.756}          & { 0.857}          & { 0.699}          & { 0.820}          \\
{ MedSAM (NC'24)\cite{ma2024segment}}            & { 0.749}          & { 0.851}          & { 0.694}          & { 0.812}          \\
{ \textbf{Ours}}                     & { \textbf{0.904}} & { \textbf{0.948}} & { \textbf{0.919}} & { \textbf{0.957}} \\ \hline
\label{tab:2}
\end{tabular}
\end{table}

\begin{figure}[!ht]
\centering
\includegraphics[width=0.49\textwidth]{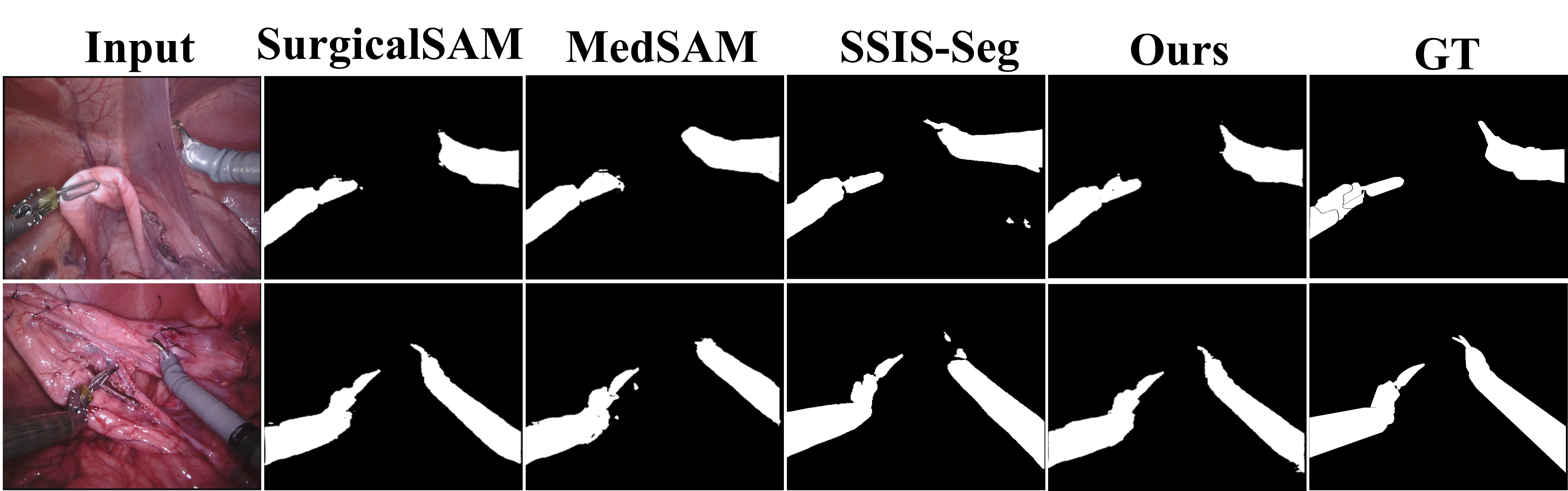}
\caption{Visualization of surgical instrument segmentation results. Qualitative results on instrument segmentation demonstrate that LapaFM produces more accurate and coherent edge predictions, supporting enhanced intraoperative navigation.}
\label{fig: surgical_instrument_segmentation}
\end{figure}
\subsubsection{Surgical Instrument Segmentation}

For surgical instrument segmentation, a segmentation head was attached to LapaFM with the backbone frozen and fine-tuned on downstream datasets. LapaFM was compared with task-specific models (SSIS-Seg\cite{colleoni2022ssis}, Active\_Learning\cite{peng2024reducing}, and MCF-SMSIS\cite{wu2024mcf}) and generalist models (SurgicalSAM\cite{yue2024surgicalsam} and MedSAM\cite{ma2024segment}). As shown in Table\ref{tab:2}, LapaFM achieved the best performance on both EndoVis2017 and EndoVis2018, with Dice scores of 0.948 and 0.957, outperforming SSIS-Seg by 12.1\% and 17.3\%, respectively. Consistent improvements were also observed over SurgicalSAM (10.6\% and 16.7\%) and MedSAM (11.3\% and 17.8\%). Qualitative results in Fig.\ref{fig: surgical_instrument_segmentation} further demonstrate that LapaFM achieves more accurate localization and segmentation by effectively capturing semantic features of surgical instruments.

\subsection{LapaFM supports the prediction of lymph node metastasis in colorectal cancer}

Intraoperative lesion identification and classification are crucial for surgical accuracy and patient prognosis. As an essential component of modern precision surgery, intraoperative identification supports surgical decision-making, improves surgical precision and safety, and ultimately enhances patient outcomes. Based on this clinical need, we designed a downstream task for colorectal cancer lymph node metastasis (LNM) prediction. The experiments were conducted on the CRCLNM dataset, which contains 678 training images, 281 validation images, and 270 test images.

To evaluate the diagnostic capability of LapaFM, we compared its performance in LNM prediction with several specialized classification models, including Swin-Transformer\cite{liu2021swin}, ConvNeXt\cite{liu2022convnet}, EfficientNetv2\cite{tan2021efficientnetv2}, and MFI-FFP\cite{zhu2024prediction}, as well as large-scale foundation models such as GastroNet\_5M\cite{jong2025gastronet} and DINOv3\cite{simeoni2025dinov3}. The experimental results demonstrate that LapaFM consistently outperforms these models, highlighting its superior effectiveness in the lymph node metastasis prediction task.

\begin{figure}[!ht]
\centering
\includegraphics[width=0.49\textwidth]{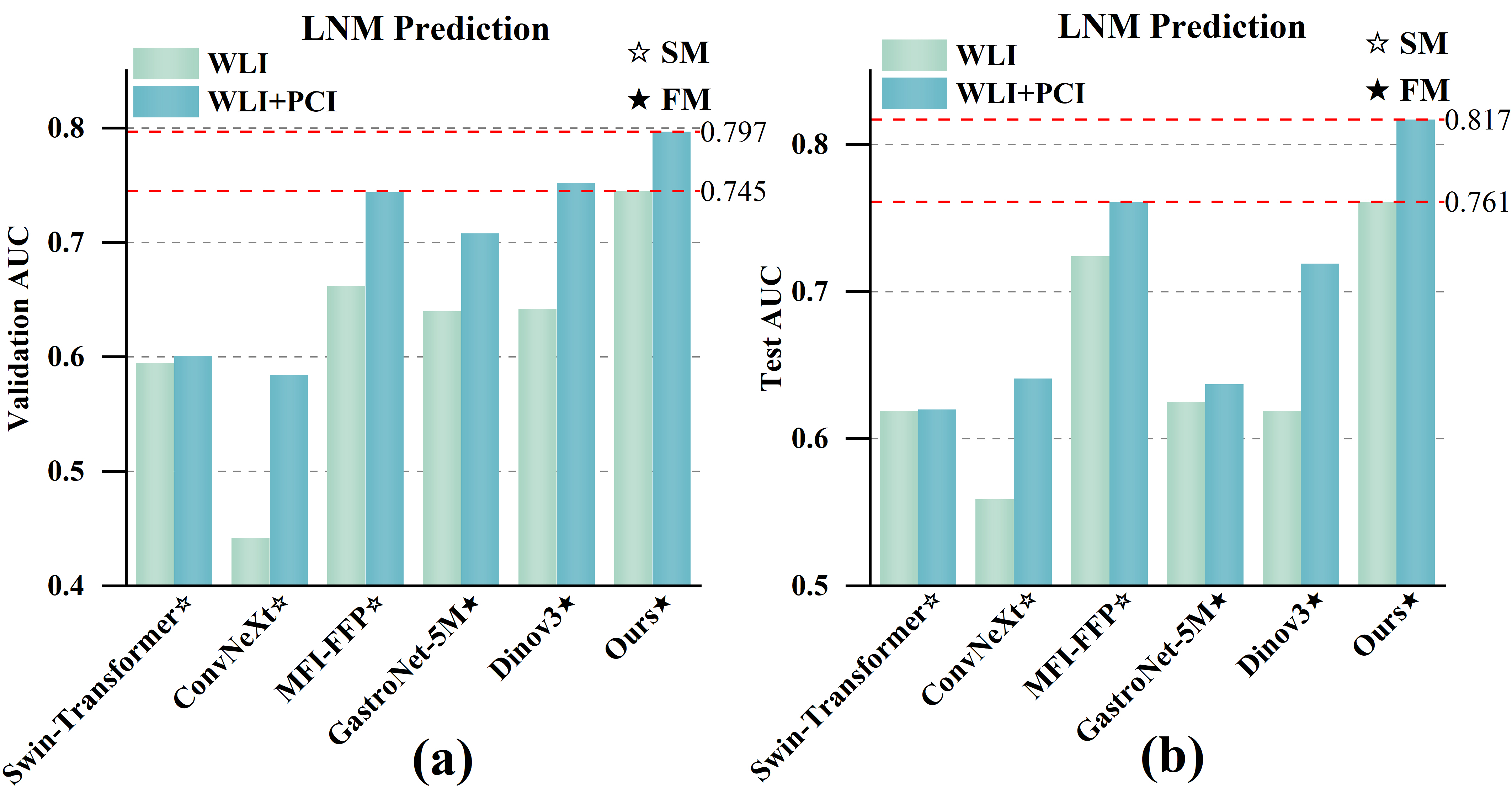}
\caption{Comparison of results for the task of predicting lymph node metastasis in colorectal cancer. a. Performance comparison of LapaFM on the validation set for the colorectal cancer lymph node metastasis prediction task. b. Performance comparison of LapaFM on the test set for the colorectal cancer lymph node metastasis prediction task.}
\label{fig: LNM_prediction}
\end{figure}
The performance of LapaFM on the LNM prediction task is illustrated in Figs.\ref{fig: LNM_prediction}a and \ref{fig: LNM_prediction}b. When trained using only unimodal white-light images, LapaFM achieves AUC values of 0.745 and 0.761 on the validation and test sets, respectively. These results represent improvements of 8.3\% and 3.7\% compared with the task-specific MFI-FFP method. Furthermore, compared with other foundation models such as GastroNet-5M and DINOv3, LapaFM achieves state-of-the-art performance under the unimodal setting, underscoring the critical role of high-quality pretraining data in enhancing downstream generalization capability.

When both white-light and pseudo-color modalities were used, LapaFM further improved performance, achieving AUCs of 0.797 and 0.817 on the validation and test sets, respectively. This demonstrates the effectiveness of the multi-stage incremental learning strategy in integrating complementary modality-specific knowledge. 

\begin{table}[!htbp]
\renewcommand\arraystretch{1.3}
\setlength\tabcolsep{10pt} 
\caption{Comparison between multi-stage continuous learning and joint training for lymph node metastasis prediction, demonstrating improved multimodal modeling with reduced modality conflict and knowledge loss.}
\centering
\begin{tabular}{ccc}
\hline
\multirow{2}{*}{\textbf{Pretraining Strategy}} & \textbf{Validation Set} & \textbf{Test Set}  \\ \cline{2-3} 
                          & \textbf{AUC}        & \textbf{AUC}   \\ \hline
Joint Pretraining                     & 0.783      & 0.722 \\
\textbf{Ours}                      & \textbf{0.797 }     & 0.\textbf{817} \\ \hline
\label{tab:3}
\end{tabular}
\end{table}

\begin{table*}[!htbp]
\renewcommand\arraystretch{1.3}
\setlength\tabcolsep{12pt} 
\caption{Quantitative results on surgical instrument segmentation show that LapaFM outperforms three task-specific models and two general-purpose segmentation models, achieving the highest IoU and Dice scores.}
\centering
\begin{tabular}{ccccccccc}
\hline
                                & \multicolumn{2}{c}{{ \textbf{CVC-ClinicDB}}}           & \multicolumn{2}{c}{{ \textbf{CVC-ColonDB}}}            & \multicolumn{2}{c}{{ \textbf{ETIS}}}                   & \multicolumn{2}{c}{{ \textbf{EndoScene}}}              \\ \cline{2-9} 
\multirow{-2}{*}{{ \textbf{Model}}} & { \textbf{IOU}} & { \textbf{DICE}} & { \textbf{IOU}} & { \textbf{DICE}} & { \textbf{IOU}} & { \textbf{DICE}} & { \textbf{IOU}} & { \textbf{DICE}} \\ \hline
{ Polyper (AAAI'2024)\cite{shao2024polyper}}              & { 0.889}        & { 0.942}         & { 0.668}        & { 0.756}         & { 0.785}        & { 0.865}         & { 0.867}        & { 0.924}         \\
{ CFANet (PR'2023)\cite{zhou2023cross}}                 & { 0.883}        & { 0.933}         & { 0.665}        & { 0.743}         & { 0.655}        & { 0.732}         & { 0.827}        & { 0.893}         \\
{ CoInNet (TMI'2023)\cite{jain2023coinnet}}               & { 0.887}        & { 0.930}         & { 0.729}        & { 0.797}         & { 0.690}        & { 0.759}         & { 0.863}        & { 0.909}         \\
{ BUNet (NN'2023)\cite{yue2024boundary}}                  & { 0.890}        & { 0.935}         & { 0.731}        & { 0.813}         & { 0.721}        & { 0.798}         & { 0.834}        & { 0.900}         \\
{ SurgicalSAM (AAAI'2024)\cite{yue2024surgicalsam}}          & { 0.505}        & { 0.644}         & { 0.330}        & { 0.460}         & { 0.238}        & { 0.342}         & { 0.472}        & { 0.623}         \\
{ MedSAM (NC'2024)\cite{ma2024segment}}                 & { 0.803}        & { 0.867}         & { 0.651}        & { 0.734}         & { 0.604}        & { 0.687}         & { 0.798}        & { 0.870}         \\
\textbf{Ours}                                           & \textbf{0.896}                      & \textbf{0.943}                       & \textbf{0.743}                      & \textbf{0.835}                       & \textbf{0.796}                      & \textbf{0.875}                       & \textbf{0.877}                      & \textbf{0.931}                       \\ \hline
\label{tab:4}
\end{tabular}
\end{table*}
\begin{table*}[!htbp]
\renewcommand\arraystretch{1.3}
\setlength\tabcolsep{10pt} 
\caption{Comparison of LapaFM performance on polyp detection and classification across the CP-CHILD and WCE-CCD datasets.}
\centering
\begin{tabular}{ccccccc}
\hline
\multirow{3}{*}{\textbf{Model}} & \multicolumn{2}{c}{\textbf{Polyp Detection}} & \multicolumn{4}{c}{\textbf{Polyp  Classification}}                                                  \\ \cline{2-7} 
                                & \multicolumn{2}{c}{\textbf{CH-CHILD}}        & \multicolumn{2}{c}{\textbf{WCE-CCD validation set}} & \multicolumn{2}{c}{\textbf{WCE-CCD test set}} \\ \cline{2-7} 
                                & \textbf{ACC}          & \textbf{AUC}         & \textbf{ACC}             & \textbf{F1}              & \textbf{ACC}          & \textbf{F1}           \\ \hline
Swin-Transformer (ICCV'21)\cite{liu2021swin}               & 0.956                 & 0.992                & 0.827                    & 0.821                    & 0.831                 & 0.828                 \\
ConvNeXt (CVPR'22)\cite{liu2022convnet}                      & 0.979                 & 0.992                & 0.761                    & 0.741                    & 0.795                 & 0.780                 \\
Efficientnetv2 (ICML'21)\cite{tan2021efficientnetv2}                & 0.961                 & 0.994                & 0.772                    & 0.766                    & 0.792                 & 0.782                 \\
GastroNet\_5M (GAS'25)\cite{jong2025gastronet}                   & 0.980                 & 0.996                & 0.941                    & 0.932                    & 0.957                 & 0.935                 \\
Dinov3 (Arxiv'25)\cite{simeoni2025dinov3}                         & 0.987                 & 0.998                & 0.912                    & 0.902                    & 0.926                 & 0.911                 \\
\textbf{Ours}                   & \textbf{0.992}        & \textbf{0.999}       & \textbf{0.968}           & \textbf{0.953}           & \textbf{0.971}        & \textbf{0.957}        \\ \hline
\label{tab:5}
\end{tabular}
\end{table*}
\subsection{Ablation Experiment}
To validate the proposed training strategy, LapaFM was compared with a jointly trained baseline in which white-light and pseudo-color data are learned simultaneously. As shown in Table\ref{tab:3}, LapaFM consistently outperforms the baseline on both validation and test sets. This improvement demonstrates that the multi-stage incremental pretraining strategy effectively preserves modality-specific representations while progressively integrating complementary information. In particular, it facilitates the incorporation of pseudo-color (fluorescence) features while retaining essential white-light knowledge, thereby reducing modality interference and mitigating catastrophic forgetting. These results underscore the importance of staged learning for stable and effective multimodal representation in intraoperative imaging.


\subsection{Model generalization ability analysis and prospective validation}
To evaluate the generalization capability of LapaFM, we further extended it to gastrointestinal endoscopy tasks, including polyp segmentation, detection, and classification. Moreover, we  compared LapaFM against several specialized models (SM), including Polyper\cite{shao2024polyper}, CFANet\cite{zhou2023cross}, CoInNet\cite{jain2023coinnet}, and BUNet\cite{yue2024boundary} for polyp segmentation. Compared its performance with Swin-Transformer\cite{liu2021swin}, ConvNeXt\cite{liu2022convnet}, and EfficientNetv2\cite{tan2021efficientnetv2}, as well as large-scale foundation models such as GastroNet\_5M\cite{jong2025gastronet}  and DINOv3\cite{simeoni2025dinov3} for polyp detection and classification. The experimental results demonstrate that LapaFM consistently outperforms these models, highlighting its strong generalization capability across both tasks.

\subsubsection{polyp segmentation}
As illustrated in Table \ref{tab:4}, LapaFM achieved the best performance across all four test datasets. Compared to the specialized model Polyper, LapaFM improved Dice scores by 10.4\% on CVC-ColonDB and 0.76\% on ETIS, demonstrating superior robustness and segmentation accuracy across different datasets. Furthermore, relative to the base model MedSAM, LapaFM achieved Dice improvements of 8.8\%, 13.7\%, 21.5\%, and 7.0\% on CVC-ClinicDB, CVC-ColonDB, ETIS, and EndoScene, respectively.
\subsubsection{polyp detection and classification}
As illustrated in Table \ref{tab:5}, LapaFM achieves outstanding performance in the polyp detection task, reaching an AUC of 0.999, surpassing all foundation models and specialized models. This result indicates that LapaFM can effectively identify the presence of intra-abdominal polyps, demonstrating high reliability in lesion classification within gastrointestinal endoscopy.

Furthermore, as shown in Table \ref{tab:5}, LapaFM captures tissue-specific features more effectively and exhibits stronger generalization and robustness compared with other models. It achieves F1 scores of 0.953 and 0.957 on the WCE-CCD validation and test sets, respectively. These results further confirm the superior transferability and stability of LapaFM in gastrointestinal endoscopic analysis tasks.

\begin{table}[!htbp]
\renewcommand\arraystretch{1.3}
\setlength\tabcolsep{4pt} 
\caption{Performance comparison of LapaFM on prospective polyp classification and detection tasks across Propolyp-cls and Propolyp-detection dataset.}
\centering
\begin{tabular}{ccccc}
\hline
\multirow{2}{*}{\textbf{Model}} & \multicolumn{2}{c}{\textbf{Propolyp-cls}} & \multicolumn{2}{c}{\textbf{Propolyp-detection}} \\ \cline{2-5} 
                                & \textbf{ACC}         & \textbf{AUC}       & \textbf{ACC}           & \textbf{AUC}           \\ \hline
Swin-Transformer(ICCV'21)\cite{liu2021swin}                  & 0.646                & 0.535              & 0.554                  & 0.523                  \\
ConvNeXt (CVPR'22)\cite{liu2022convnet}                       & 0.667                & 0.562              & 0.593                  & 0.529                  \\
Efficientnetv2(ICML'21)\cite{tan2021efficientnetv2}                  & 0.671                & 0.576              & 0.619                  & 0.645                  \\
GastroNet\_5M (GAS'25)\cite{jong2025gastronet}                     & 0.687                & 0.594              & 0.796                  & 0.81                   \\
Dinov3 (Arxiv'25)\cite{simeoni2025dinov3}                          & 0.675                & 0.582              & 0.787                  & 0.779                  \\
\textbf{ours}                   & \textbf{0.738}       & \textbf{0.69}      & \textbf{0.823}         & \textbf{0.838}         \\ \hline
\label{tab:6}
\end{tabular}
\end{table}
\subsubsection{prospective validation}
To further assess generalization, we evaluated LapaFM on a prospective polyp dataset derived from the clinical trial NCT06846229, comprising 300 polyp images from 115 patients (40 females and 75 males; age range: 29–86 years), including adenomas and hyperplastic polyps. The dataset was split into 100 training and 200 testing images. As shown in Table \ref{tab:6}, LapaFM achieved the best performance among all compared models, although the AUC (0.690) was lower than that on other datasets, likely due to the domain gap between laparoscopic pretraining data and colonoscopy-specific polyp characteristics. Nevertheless, LapaFM still outperformed other foundation models, demonstrating strong cross-domain generalization.

For prospective polyp detection, the same 300 polyp images were used, together with 100 non-polyp frames randomly sampled from colonoscopy videos without preprocessing, introducing substantial motion, illumination, and fluid artifacts. As shown in Table \ref{tab:6}, LapaFM achieved an AUC of 0.838, outperforming all competing methods. These results indicate that LapaFM maintains robust detection performance under noisy conditions and exhibits strong generalization to previously unseen clinical data.

\begin{figure}[!ht]
\centering
\includegraphics[width=0.5\textwidth]{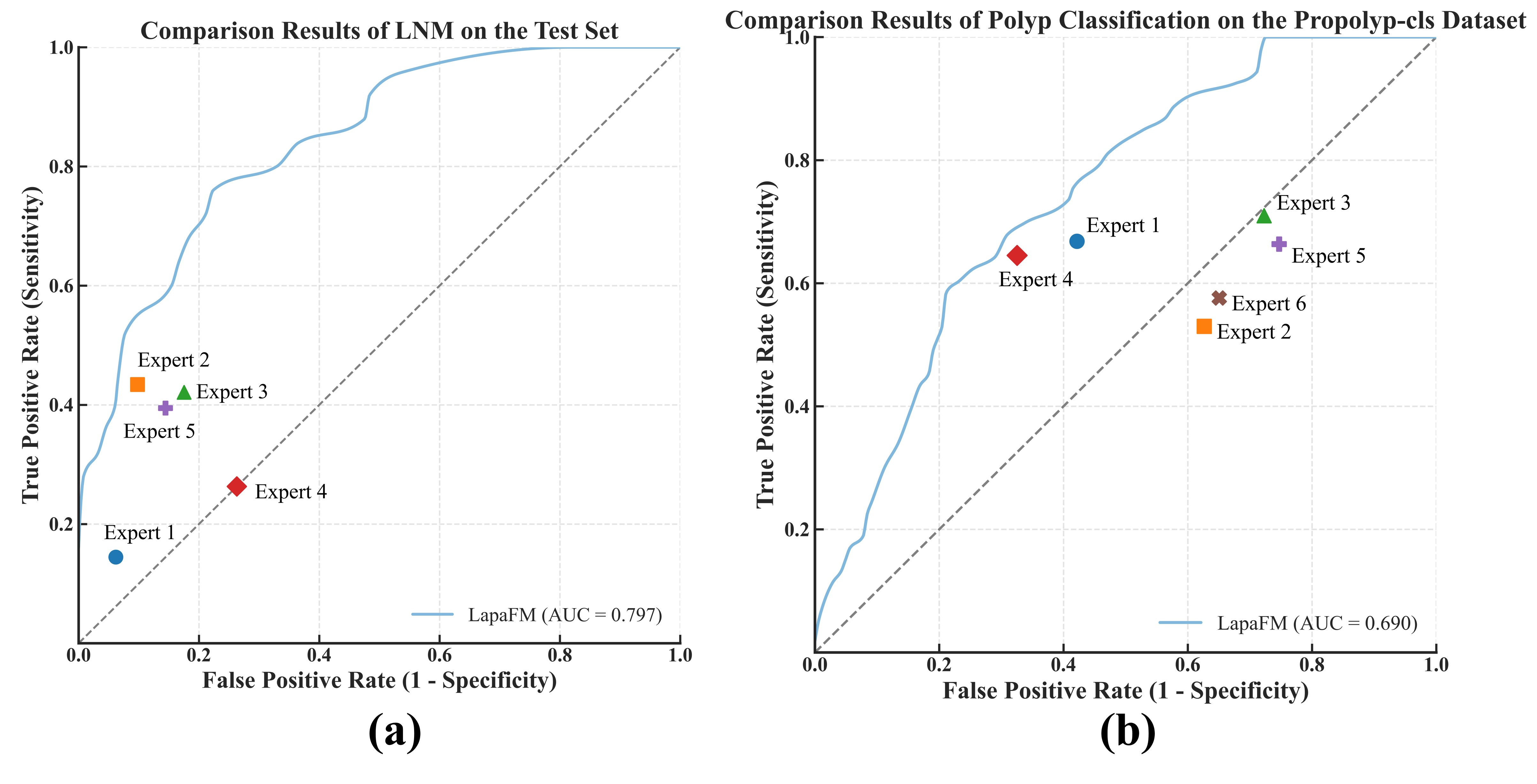}
\caption{Comparison between LapaFM and clinical experts. a. Lymph node metastasis prediction across five gastroenterologists with varying experience levels. b. Polyp classification across six endoscopy experts with varying experience levels.}
\label{fig: Clinical_Experts}
\end{figure}
\subsection{LapaFM Compared with Clinical Experts}
To evaluate the clinical relevance of LapaFM, we compared its performance with that of experienced clinicians on gastrointestinal tasks. In lymph node metastasis prediction, LapaFM outperformed five gastroenterologists with varying levels of experience (Fig.\ref{fig: Clinical_Experts}a), demonstrating superior diagnostic accuracy and potential to assist surgical decision-making. In polyp classification on a prospective dataset, six endoscopy experts with different experience levels participated in the evaluation (Fig.\ref{fig: Clinical_Experts}b). The results showed that differentiating hyperplastic polyps from adenomas remains challenging even for experienced clinicians, whereas LapaFM achieved consistently higher accuracy. These findings highlight the strong clinical potential of LapaFM in supporting expert-level decision-making across diverse gastrointestinal tasks.

\subsection{Model Robustness Analysis}
Robustness is fundamental to diagnostic accuracy, patient safety, and the clinical reliability of medical foundation models. It determines whether AI can be safely deployed in dynamic, high-stakes surgical environments. As AI advances toward applications in personalized medicine and intraoperative guidance, robustness remains essential for safe clinical translation. To evaluate the robustness of LapaFM, we tested its performance under varying training data scales across multiple downstream tasks.

\begin{figure}[!ht]
\centering
\includegraphics[width=0.49\textwidth]{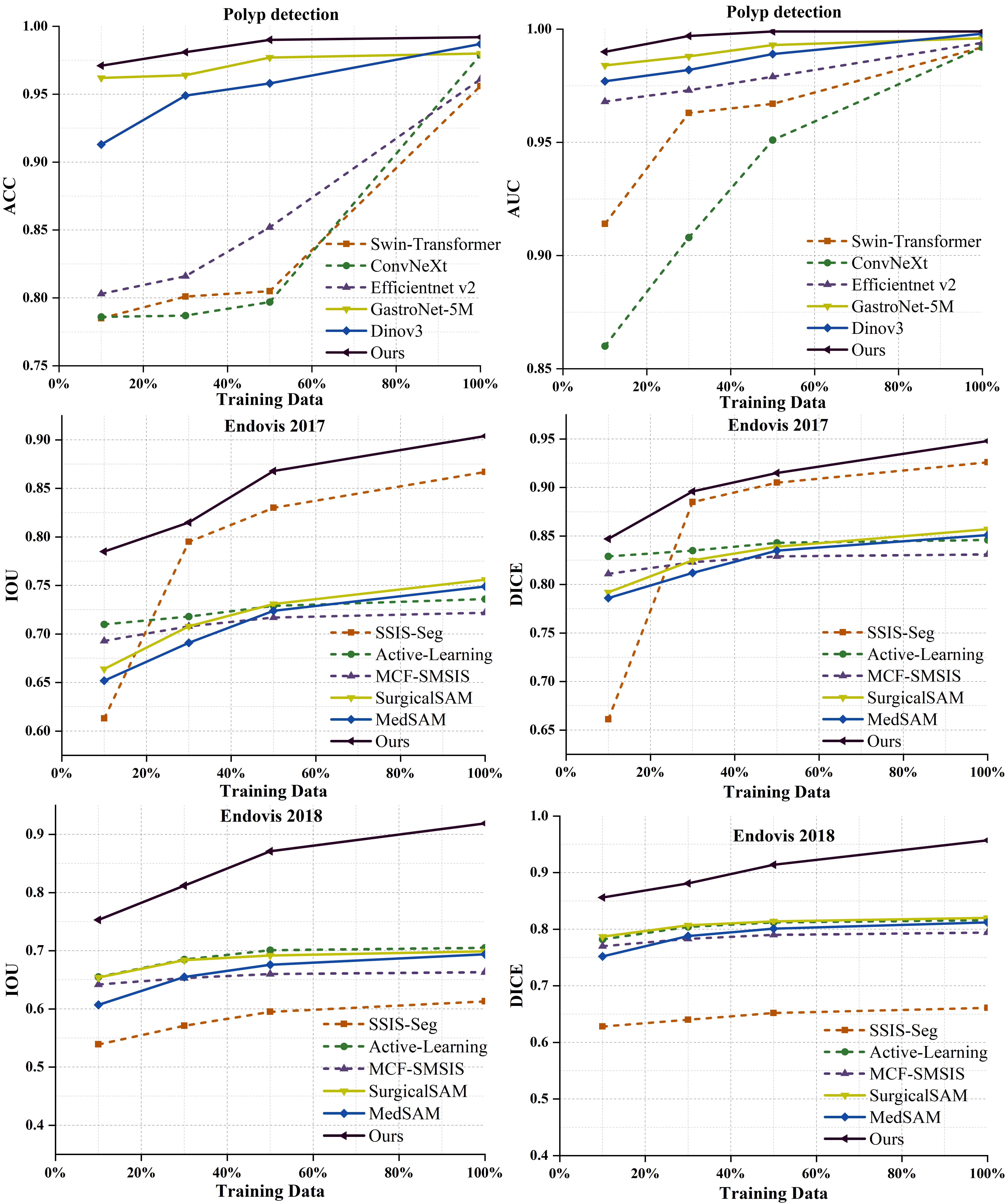}
\caption{Robustness of LapaFM under limited supervision. Performance on instrument segmentation and polyp detection with 10\%, 30\%, and 50\% training data demonstrates superior robustness compared to competing models.}
\label{fig: Robustness_Analysis}
\end{figure}
As illustrated in Fig.\ref{fig: Robustness_Analysis}, in the polyp detection task, models trained with 10\%, 30\%, and 50\% of the data consistently achieved AUCs $>$0.990, surpassing all competing approaches and demonstrating strong classification stability. For surgical instrument segmentation, models trained on 10–50\% of the EndoVis2017 and EndoVis2018 datasets maintained Dice scores above 0.847 and 0.850, respectively. In contrast, task-specific and general segmentation frameworks such as SurgicalSAM and MedSAM showed marked performance degradation with reduced data, particularly on the more complex EndoVis2018 dataset. These findings highlight LapaFM’s superior generalization and robustness across diverse surgical tasks and data regimes, underscoring its potential for reliable deployment in real-world laparoscopic settings with limited labeled data.

\begin{figure}[!ht]
\centering
\includegraphics[width=0.5\textwidth]{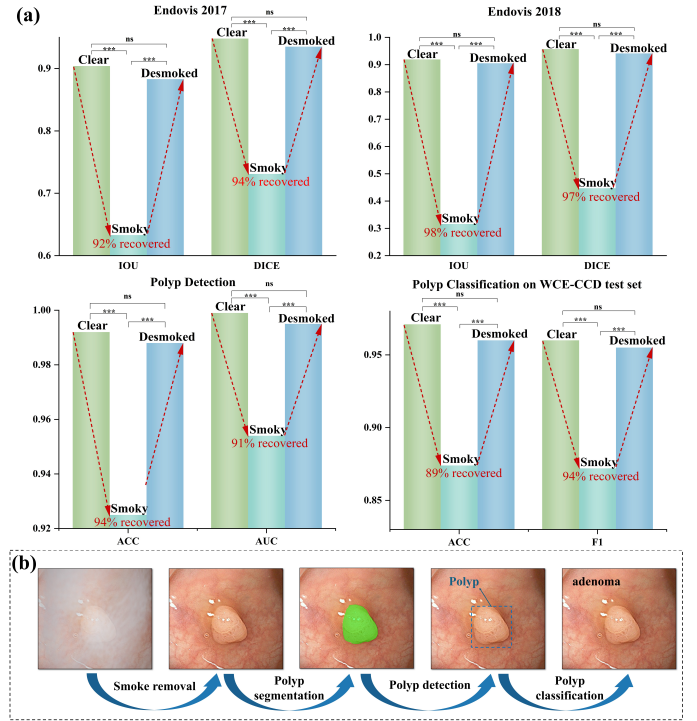}
\caption{Effect of surgical smoke on downstream tasks and full-process surgical workflow support. a. Surgical smoke severely degrades segmentation and classification performance by obscuring the visual field. After applying LapaFM-based smoke removal, performance is significantly improved and approaches that of clear images, with no significant difference(ns, 
$P>0.05$; $***$, $P<0.001$), indicating effective recovery of critical visual information. b. LapaFM supports the full laparoscopic workflow, from intraoperative smoke removal to downstream lesion classification.}
\label{fig: Surgical_Workflow}
\end{figure}
\subsection{LapaFM Assisted Surgical Workflow}
To further evaluate the impact of surgical smoke on intraoperative tasks, we introduced synthetic smoke into downstream datasets and assessed performance before and after smoke removal using LapaFM. As shown in Fig.\ref{fig: Surgical_Workflow}a, segmentation performance on EndoVis2017 and EndoVis2018 improved substantially after smoke removal, with Dice scores reaching 0.935 and 0.941, respectively, showing no significant difference compared to smoke-free conditions ($P>0.05$), indicating effective recovery of visual information. For polyp detection, the AUC increased to 0.995, closely matching the baseline on clear images. In contrast, polyp classification performance was significantly degraded by smoke ($P<0.001$) and only partially recovered after desmoking, likely due to residual artifacts and loss of fine morphological details. These results highlight both the severe adverse effects of smoke on intraoperative perception and the strong desmoking capability of LapaFM. Furthermore, LapaFM enables an integrated AI-assisted surgical workflow (Fig.\ref{fig: Surgical_Workflow}b), encompassing smoke removal, instrument and anatomical segmentation, and downstream diagnostic tasks such as polyp detection, classification, and lymph node metastasis prediction. This unified framework enhances intraoperative visualization, localization, and decision-making, demonstrating strong robustness, generalization, and potential for improving the safety and efficiency of laparoscopic surgery.

\section{DISCUSSION AND CONCLUSION}
In this work, we propose LapaFM, a multimodal intraoperative imaging foundation model tailored for laparoscopic surgery, which integrates white-light and pseudo-color modalities through large-scale pretraining on 250 surgical videos and a multi-stage incremental distillation strategy to mitigate catastrophic forgetting. Extensive evaluations across diverse downstream tasks, including smoke removal, surgical phase recognition, instrument and polyp segmentation, polyp detection and classification, and lymph node metastasis prediction, demonstrate its effectiveness, with additional prospective validation confirming its clinical applicability. The strong performance of LapaFM is attributed to its ability to learn hierarchical representations from large-scale heterogeneous data and to achieve stable multimodal fusion via continual distillation, enabling robust generalization across classification, detection, and segmentation tasks. Moreover, the model exhibits strong robustness and data efficiency, maintaining high performance under limited supervision and generalizing well to unseen clinical scenarios. Despite these advantages, limitations remain, including restricted modality coverage, limited temporal modeling capability, and deployment challenges due to model scale. Future work will focus on expanding multimodal inputs, improving spatiotemporal modeling for long surgical videos, and conducting large-scale prospective validation. Overall, LapaFM provides a unified and scalable framework for addressing key intraoperative challenges and advancing surgical intelligence toward safer and more effective laparoscopic procedures.

\bibliographystyle{IEEEtran}
\bibliography{ref} 

@article{bonjer2015randomized,
  title={A randomized trial of laparoscopic versus open surgery for rectal cancer},
  author={Bonjer, H Jaap and Deijen, Charlotte L and Abis, Gabor A and Cuesta, Miguel A and Van Der Pas, Martijn HGM and De Lange-De Klerk, Elly SM and Lacy, Antonio M and Bemelman, Willem A and Andersson, John and Angenete, Eva and others},
  journal={New England Journal of Medicine},
  volume={372},
  number={14},
  pages={1324--1332},
  year={2015},
  publisher={Mass Medical Soc}
}

@article{wang2023fluorescence,
  title={Fluorescence image-guided tumour surgery},
  author={Wang, Kun and Du, Yang and Zhang, Zeyu and He, Kunshan and Cheng, Zhongquan and Yin, Lin and Dong, Di and Li, Changjian and Li, Wei and Hu, Zhenhua and others},
  journal={Nature Reviews Bioengineering},
  volume={1},
  number={3},
  pages={161--179},
  year={2023},
  publisher={Nature Publishing Group UK London}
}

@article{feng2022development,
  title={Development and validation of a radiopathomics model to predict pathological complete response to neoadjuvant chemoradiotherapy in locally advanced rectal cancer: a multicentre observational study},
  author={Feng, Lili and Liu, Zhenyu and Li, Chaofeng and Li, Zhenhui and Lou, Xiaoying and Shao, Lizhi and Wang, Yunlong and Huang, Yan and Chen, Haiyang and Pang, Xiaolin and others},
  journal={The Lancet Digital Health},
  volume={4},
  number={1},
  pages={e8--e17},
  year={2022},
  publisher={Elsevier}
}

@article{moor2023foundation,
  title={Foundation models for generalist medical artificial intelligence},
  author={Moor, Michael and Banerjee, Oishi and Abad, Zahra Shakeri Hossein and Krumholz, Harlan M and Leskovec, Jure and Topol, Eric J and Rajpurkar, Pranav},
  journal={Nature},
  volume={616},
  number={7956},
  pages={259--265},
  year={2023},
  publisher={Nature Publishing Group UK London}
}

@article{song2023vision,
  title={Vision transformers for single image dehazing},
  author={Song, Yuda and He, Zhuqing and Qian, Hui and Du, Xin},
  journal={IEEE Transactions on Image Processing},
  volume={32},
  pages={1927--1941},
  year={2023},
  publisher={IEEE}
}

@article{xia2025vivo,
  title={In Vivo Laparoscopic Image De-Smoking Dataset, Evaluation, and Beyond},
  author={Xia, Wenyao and Peters, Terry M and Fan, Victoria and Sthanunathan, Hamsini and Qi, Olivia and Chen, Elvis CS},
  journal={IEEE Transactions on Medical Imaging},
  year={2025},
  publisher={IEEE}
}

@article{wang2020improved,
  title={An improved deep learning approach and its applications on colonic polyp images detection},
  author={Wang, Wei and Tian, Jinge and Zhang, Chengwen and Luo, Yanhong and Wang, Xin and Li, Ji},
  journal={BMC Medical Imaging},
  volume={20},
  pages={1--14},
  year={2020},
  publisher={Springer}
}

@article{montalbo2022diagnosing,
  title={Diagnosing gastrointestinal diseases from endoscopy images through a multi-fused CNN with auxiliary layers, alpha dropouts, and a fusion residual block},
  author={Montalbo, Francis Jesmar P},
  journal={Biomedical signal processing and control},
  volume={76},
  pages={103683},
  year={2022},
  publisher={Elsevier}
}

@article{zhu2024prediction,
  title={Prediction of lymph node metastasis in colorectal cancer using intraoperative fluorescence multi-modal imaging},
  author={Zhu, Xiaobo and Sun, He and Wang, Yuhan and Hu, Gang and Shao, Lizhi and Zhang, Song and Liu, Fucheng and Chi, Chongwei and He, Kunshan and Tang, Jianqiang and others},
  journal={IEEE Transactions on Medical Imaging},
  volume={44},
  number={3},
  pages={1568--1580},
  year={2024},
  publisher={IEEE}
}

@article{allan20192017,
  title={2017 robotic instrument segmentation challenge},
  author={Allan, Max and Shvets, Alex and Kurmann, Thomas and Zhang, Zichen and Duggal, Rahul and Su, Yun-Hsuan and Rieke, Nicola and Laina, Iro and Kalavakonda, Niveditha and Bodenstedt, Sebastian and others},
  journal={arXiv preprint arXiv:1902.06426},
  year={2019}
}

@article{allan20202018,
  title={2018 robotic scene segmentation challenge},
  author={Allan, Max and Kondo, Satoshi and Bodenstedt, Sebastian and Leger, Stefan and Kadkhodamohammadi, Rahim and Luengo, Imanol and Fuentes, Felix and Flouty, Evangello and Mohammed, Ahmed and Pedersen, Marius and others},
  journal={arXiv preprint arXiv:2001.11190},
  year={2020}
}

@inproceedings{fan2020pranet,
  title={Pranet: Parallel reverse attention network for polyp segmentation},
  author={Fan, Deng-Ping and Ji, Ge-Peng and Zhou, Tao and Chen, Geng and Fu, Huazhu and Shen, Jianbing and Shao, Ling},
  booktitle={International conference on medical image computing and computer-assisted intervention},
  pages={263--273},
  year={2020},
  organization={Springer}
}

@inproceedings{wang2022autolaparo,
  title={Autolaparo: A new dataset of integrated multi-tasks for image-guided surgical automation in laparoscopic hysterectomy},
  author={Wang, Ziyi and Lu, Bo and Long, Yonghao and Zhong, Fangxun and Cheung, Tak-Hong and Dou, Qi and Liu, Yunhui},
  booktitle={International Conference on Medical Image Computing and Computer-Assisted Intervention},
  pages={486--496},
  year={2022},
  organization={Springer}
}

@article{twinanda2016endonet,
  title={Endonet: a deep architecture for recognition tasks on laparoscopic videos},
  author={Twinanda, Andru P and Shehata, Sherif and Mutter, Didier and Marescaux, Jacques and De Mathelin, Michel and Padoy, Nicolas},
  journal={IEEE transactions on medical imaging},
  volume={36},
  number={1},
  pages={86--97},
  year={2016},
  publisher={IEEE}
}

@article{hong2023mars,
  title={MARS-GAN: multilevel-feature-learning attention-aware based generative adversarial network for removing surgical smoke},
  author={Hong, Tingxuan and Huang, Pu and Zhai, Xiangyu and Gu, Changming and Tian, Baolong and Jin, Bin and Li, Dengwang},
  journal={IEEE Transactions on Medical Imaging},
  volume={42},
  number={8},
  pages={2299--2312},
  year={2023},
  publisher={IEEE}
}

@inproceedings{wu2021contrastive,
  title={Contrastive learning for compact single image dehazing},
  author={Wu, Haiyan and Qu, Yanyun and Lin, Shaohui and Zhou, Jian and Qiao, Ruizhi and Zhang, Zhizhong and Xie, Yuan and Ma, Lizhuang},
  booktitle={Proceedings of the IEEE/CVF conference on computer vision and pattern recognition},
  pages={10551--10560},
  year={2021}
}

@article{zhou2022synchronizing,
  title={Synchronizing detection and removal of smoke in endoscopic images with cyclic consistency adversarial nets},
  author={Zhou, Yichao and Hu, Zhisen and Xuan, Zuxing and Wang, Yangang and Hu, Xiyuan},
  journal={IEEE/ACM Transactions on Computational Biology and Bioinformatics},
  volume={21},
  number={4},
  pages={670--680},
  year={2022},
  publisher={IEEE}
}

@article{venkatesh2020unsupervised,
  title={Unsupervised smoke to desmoked laparoscopic surgery images using contrast driven Cyclic-DesmokeGAN},
  author={Venkatesh, Vishal and Sharma, Neeraj and Srivastava, Vivek and Singh, Munendra},
  journal={Computers in Biology and Medicine},
  volume={123},
  pages={103873},
  year={2020},
  publisher={Elsevier}
}

@article{chen2024dea,
  title={DEA-Net: Single image dehazing based on detail-enhanced convolution and content-guided attention},
  author={Chen, Zixuan and He, Zewei and Lu, Zhe-Ming},
  journal={IEEE transactions on image processing},
  volume={33},
  pages={1002--1015},
  year={2024},
  publisher={IEEE}
}

@article{jin2025mb,
  title={MB-TaylorFormer V2: Improved multi-branch linear transformer expanded by Taylor formula for image restoration},
  author={Jin, Zhi and Qiu, Yuwei and Zhang, Kaihao and Li, Hongdong and Luo, Wenhan},
  journal={IEEE Transactions on Pattern Analysis and Machine Intelligence},
  year={2025},
  publisher={IEEE}
}

@inproceedings{fang2025guided,
  title={Guided real image dehazing using ycbcr color space},
  author={Fang, Wenxuan and Fan, Junkai and Zheng, Yu and Weng, Jiangwei and Tai, Ying and Li, Jun},
  booktitle={Proceedings of the AAAI conference on artificial intelligence},
  volume={39},
  number={3},
  pages={2906--2914},
  year={2025}
}

@inproceedings{guo2025mambairv2,
  title={Mambairv2: Attentive state space restoration},
  author={Guo, Hang and Guo, Yong and Zha, Yaohua and Zhang, Yulun and Li, Wenbo and Dai, Tao and Xia, Shu-Tao and Li, Yawei},
  booktitle={Proceedings of the Computer Vision and Pattern Recognition Conference},
  pages={28124--28133},
  year={2025}
}

@article{gaurav2005ciede2000,
  title={The CIEDE2000 color-difference formula: Implementation notes, supplementary test data, and mathematical observations},
  author={Gaurav, Sharma},
  journal={COLOR research and application},
  volume={30},
  number={1},
  pages={21--30},
  year={2005}
}

@article{ahmed2024deep,
  title={Deep learning for surgical instrument recognition and segmentation in robotic-assisted surgeries: a systematic review},
  author={Ahmed, Fatimaelzahraa Ali and Yousef, Mahmoud and Ahmed, Mariam Ali and Ali, Hasan Omar and Mahboob, Anns and Ali, Hazrat and Shah, Zubair and Aboumarzouk, Omar and Al Ansari, Abdulla and Balakrishnan, Shidin},
  journal={Artificial Intelligence Review},
  volume={58},
  number={1},
  pages={1},
  year={2024},
  publisher={Springer}
}

@article{jin2021temporal,
  title={Temporal memory relation network for workflow recognition from surgical video},
  author={Jin, Yueming and Long, Yonghao and Chen, Cheng and Zhao, Zixu and Dou, Qi and Heng, Pheng-Ann},
  journal={IEEE Transactions on Medical Imaging},
  volume={40},
  number={7},
  pages={1911--1923},
  year={2021},
  publisher={IEEE}
}

@inproceedings{czempiel2020tecno,
  title={Tecno: Surgical phase recognition with multi-stage temporal convolutional networks},
  author={Czempiel, Tobias and Paschali, Magdalini and Keicher, Matthias and Simson, Walter and Feussner, Hubertus and Kim, Seong Tae and Navab, Nassir},
  booktitle={International conference on medical image computing and computer-assisted intervention},
  pages={343--352},
  year={2020},
  organization={Springer}
}

@inproceedings{gao2021trans,
  title={Trans-svnet: Accurate phase recognition from surgical videos via hybrid embedding aggregation transformer},
  author={Gao, Xiaojie and Jin, Yueming and Long, Yonghao and Dou, Qi and Heng, Pheng-Ann},
  booktitle={International conference on medical image computing and computer-assisted intervention},
  pages={593--603},
  year={2021},
  organization={Springer}
}

@article{liu2025lovit,
  title={Lovit: Long video transformer for surgical phase recognition},
  author={Liu, Yang and Boels, Maxence and Garcia-Peraza-Herrera, Luis C and Vercauteren, Tom and Dasgupta, Prokar and Granados, Alejandro and Ourselin, Sebastien},
  journal={Medical Image Analysis},
  volume={99},
  pages={103366},
  year={2025},
  publisher={Elsevier}
}

@article{de2025scaling,
  title={Scaling up self-supervised learning for improved surgical foundation models},
  author={de Jong, Ronald and Carolus, HJ and Franciscus, HA and van Jaarsveld, Romy C and van Hillegersberg, Richard and Josien, PW and de With, Peter HN and al Khalil, Yasmina and van Der Sommen, Fons and others},
  journal={Medical Image Analysis},
  pages={103873},
  year={2025},
  publisher={Elsevier}
}

@article{jong2025gastronet,
  title={GastroNet-5M: A multicenter dataset for developing foundation models in gastrointestinal endoscopy},
  author={Jong, Martijn R and Boers, Tim GW and Fockens, Kiki N and Jukema, Jelmer B and Kusters, Carolus HJ and Jaspers, Tim JM and van Heslinga, RAH van Eijck and Slooter, Floor C and Struyvenberg, Maarten R and Bisschops, Raf and others},
  journal={Gastroenterology},
  year={2025},
  publisher={Elsevier}
}

@article{colleoni2022ssis,
  title={SSIS-Seg: Simulation-supervised image synthesis for surgical instrument segmentation},
  author={Colleoni, Emanuele and Psychogyios, Dimitris and Van Amsterdam, Beatrice and Vasconcelos, Francisco and Stoyanov, Danail},
  journal={IEEE transactions on medical imaging},
  volume={41},
  number={11},
  pages={3074--3086},
  year={2022},
  publisher={IEEE}
}

@article{peng2024reducing,
  title={Reducing annotating load: Active learning with synthetic images in surgical instrument segmentation},
  author={Peng, Haonan and Lin, Shan and King, Daniel and Su, Yun-Hsuan and Abuzeid, Waleed M and Bly, Randall A and Moe, Kris S and Hannaford, Blake},
  journal={Medical Image Analysis},
  volume={97},
  pages={103246},
  year={2024},
  publisher={Elsevier}
}

@article{wu2024mcf,
  title={Mcf-smsis: multi-tasking with complementary functions for stereo matching and surgical instrument segmentation},
  author={Wu, Renkai and He, Changyu and Liang, Pengchen and Liu, Yinghao and Huang, Yiqi and Liu, Weiping and Shu, Biao and Xu, Panlong and Chang, Qing},
  journal={Computers in Biology and Medicine},
  volume={179},
  pages={108923},
  year={2024},
  publisher={Elsevier}
}

@inproceedings{yue2024surgicalsam,
  title={Surgicalsam: Efficient class promptable surgical instrument segmentation},
  author={Yue, Wenxi and Zhang, Jing and Hu, Kun and Xia, Yong and Luo, Jiebo and Wang, Zhiyong},
  booktitle={Proceedings of the AAAI Conference on Artificial Intelligence},
  volume={38},
  number={7},
  pages={6890--6898},
  year={2024}
}

@article{ma2024segment,
  title={Segment anything in medical images},
  author={Ma, Jun and He, Yuting and Li, Feifei and Han, Lin and You, Chenyu and Wang, Bo},
  journal={Nature communications},
  volume={15},
  number={1},
  pages={654},
  year={2024},
  publisher={Nature Publishing Group UK London}
}

@inproceedings{liu2021swin,
  title={Swin transformer: Hierarchical vision transformer using shifted windows},
  author={Liu, Ze and Lin, Yutong and Cao, Yue and Hu, Han and Wei, Yixuan and Zhang, Zheng and Lin, Stephen and Guo, Baining},
  booktitle={Proceedings of the IEEE/CVF international conference on computer vision},
  pages={10012--10022},
  year={2021}
}

@inproceedings{liu2022convnet,
  title={A convnet for the 2020s},
  author={Liu, Zhuang and Mao, Hanzi and Wu, Chao-Yuan and Feichtenhofer, Christoph and Darrell, Trevor and Xie, Saining},
  booktitle={Proceedings of the IEEE/CVF conference on computer vision and pattern recognition},
  pages={11976--11986},
  year={2022}
}

@inproceedings{tan2021efficientnetv2,
  title={Efficientnetv2: Smaller models and faster training},
  author={Tan, Mingxing and Le, Quoc},
  booktitle={International conference on machine learning},
  pages={10096--10106},
  year={2021},
  organization={PMLR}
}

@article{simeoni2025dinov3,
  title={Dinov3},
  author={Sim{\'e}oni, Oriane and Vo, Huy V and Seitzer, Maximilian and Baldassarre, Federico and Oquab, Maxime and Jose, Cijo and Khalidov, Vasil and Szafraniec, Marc and Yi, Seungeun and Ramamonjisoa, Micha{\"e}l and others},
  journal={arXiv preprint arXiv:2508.10104},
  year={2025}
}

@inproceedings{shao2024polyper,
  title={Polyper: Boundary sensitive polyp segmentation},
  author={Shao, Hao and Zhang, Yang and Hou, Qibin},
  booktitle={Proceedings of the AAAI conference on artificial intelligence},
  volume={38},
  number={5},
  pages={4731--4739},
  year={2024}
}

@article{zhou2023cross,
  title={Cross-level feature aggregation network for polyp segmentation},
  author={Zhou, Tao and Zhou, Yi and He, Kelei and Gong, Chen and Yang, Jian and Fu, Huazhu and Shen, Dinggang},
  journal={Pattern Recognition},
  volume={140},
  pages={109555},
  year={2023},
  publisher={Elsevier}
}

@article{jain2023coinnet,
  title={Coinnet: A convolution-involution network with a novel statistical attention for automatic polyp segmentation},
  author={Jain, Samir and Atale, Rohan and Gupta, Anubhav and Mishra, Utkarsh and Seal, Ayan and Ojha, Aparajita and Jaworek-Korjakowska, Joanna and Krejcar, Ondrej},
  journal={IEEE Transactions on Medical Imaging},
  volume={42},
  number={12},
  pages={3987--4000},
  year={2023},
  publisher={IEEE}
}

@article{yue2024boundary,
  title={Boundary uncertainty aware network for automated polyp segmentation},
  author={Yue, Guanghui and Zhuo, Guibin and Yan, Weiqing and Zhou, Tianwei and Tang, Chang and Yang, Peng and Wang, Tianfu},
  journal={Neural Networks},
  volume={170},
  pages={390--404},
  year={2024},
  publisher={Elsevier}
}
\end{document}